\documentclass[10pt,twocolumn,letterpaper]{article}

\usepackage{iccv}
\usepackage{times}
\usepackage{epsfig}
\usepackage{graphicx}
\usepackage{amsmath}
\usepackage{amssymb}
\usepackage{subfigure}
\usepackage{caption}
\usepackage{soul}
\usepackage{authblk}
\usepackage{multirow}
\usepackage[table,xcdraw]{xcolor}

\usepackage[breaklinks=true,bookmarks=false]{hyperref}

\iccvfinalcopy 

\begin{document}

\title{OpenEDS: Open Eye Dataset }

\author[1]{Stephan J. Garbin\thanks{This work was done during internship at Facebook Reality Labs.}}
\author[2]{Yiru Shen}
\author[2]{Immo Schuetz}
\author[2]{Robert Cavin}
\author[3]{Gregory Hughes\thanks{This work was done during employment at Facebook Reality Labs.}}
\author[2]{Sachin S. Talathi\thanks{Corresponding author}}
\affil[1]{University College London}
\affil[2]{Facebook Reality Labs}
\affil[3]{Google via Adecco}


\maketitle

\begin{abstract}
We present a large scale data set, OpenEDS: Open Eye Dataset, of eye-images captured using a virtual-reality (VR) head mounted display mounted with two synchronized eye-facing cameras at a frame rate of 200 Hz under controlled illumination.
This dataset is compiled from video capture of the eye-region collected from 152 individual participants and is divided into four subsets: (i) 12,759 images with pixel-level annotations for key eye-regions: iris, pupil and sclera (ii) 252,690 unlabelled eye-images, (iii) 91,200 frames from randomly selected video sequence of 1.5 seconds in duration and (iv) 143 pairs of left and right point cloud data compiled from corneal topography of eye regions collected from a subset, 143 out of 152, participants in the study. A baseline experiment has been evaluated on OpenEDS for the task of semantic segmentation of pupil, iris, sclera and background, with the mean intersection-over-union (mIoU) of 98.3 \%. We anticipate that OpenEDS will create opportunities to researchers in the eye tracking community and the broader machine learning and computer vision community to advance the state of eye-tracking for VR applications. The dataset is available for download upon request at \href{https://research.fb.com/programs/openeds-challenge}{https://research.fb.com/programs/openeds-challenge}

\end{abstract}

\section{Introduction}
Understanding the motion and appearance of the human eye is of great importance to many scientific fields \cite{EyeTrackingBook0}. For example, gaze direction can offer information about the focus of a person's attention \cite{borji2013state}, as well as certain aspects of their physical and psychological well-being \cite{harezlak2018application}, which in turn can facilitates the research on eye tracking to aid in the study and design of how humans interact with their environment \cite{Smith2013GazeLP}.

In the context of virtual reality (VR), accurate and precise eye tracking can enable game-changing technological advances, for example, foveated rendering, a technique that exploits the sensitivity profile of the human eye to render only those parts of a virtual scene at full resolution that the user is focused on, can significantly alleviate the computational burden of VR \cite{Patney:2016:PFV:2929464.2929472}. 
Head-mounted displays (HMDs) that include gaze-contingent variable focus \cite{Kramida2016} and gaze-driven rendering of perceptually accurate focal blur \cite{Xiao:2018:DLI:3272127.3275032} promise to alleviate vergence-accommodation-conflict and increase visual realism. 
Finally, gaze-driven interaction schemes could enable novel methods of navigating and interacting with virtual environments \cite{Tanriverdi2000}. 

The success of data driven machine learning models learned directly from images (\cite{krizhevsky2012imagenet}, \cite{DBLP:journals/corr/HeZRS15}) is accompanied by the demand for datasets of sufficient size and variety to capture the distribution of natural images sufficiently for the task at hand \cite{DBLP:journals/corr/abs-1809-04729}. 
While being able to source vast collections of images from freely available online data has led to the successful creation of datasets such as ImageNet \cite{krizhevsky2012imagenet} and COCO \cite{DBLP:journals/corr/LinMBHPRDZ14}, many other research areas require special equipment and expert knowledge for data capture. For example, the creation of the KITTI dataset required synchronization of cameras alongside a laser scanner and localization equipment (\cite{geiger2012we,geiger2013vision}. We seek to address this challenge for eye-tracking with HMDs.

Commercially available eye tracking systems often do not expose the raw camera image to the user, complicating the generation of large-scale eye image datasets. We opt to capture eye-images using a custom-built VR HMD with two synchronized cameras operating at 200Hz under controlled illumination. As outlined below, we also use specialist medical equipment to capture further information about the shape and optical properties of each participant's eyes contained in the dataset. It is the unique combination of advanced data capture, usually only performed in a clinical setting, with high resolution eye images and corresponding annotation masks for key eye-regions that sets OpenEDS apart from comparable datasets. We hope that OpenEDS bridges the gap between the vision and eye tracking communities and provides novel opportunities for research in the domain of eye-tracking for HMDs.

Our contributions are summarized as follows:
\begin{itemize}
\item A large scale dataset captured using an HMD with two synchronized cameras under controlled illumination and high frame rates;
\item A large scale dataset of annotation masks for key eye-regions: the iris, the sclera and the pupil;
\item point cloud data from corneal topography captures of eye regions.
\end{itemize}

\begin{figure}
    \centering
    \includegraphics[width=4cm]{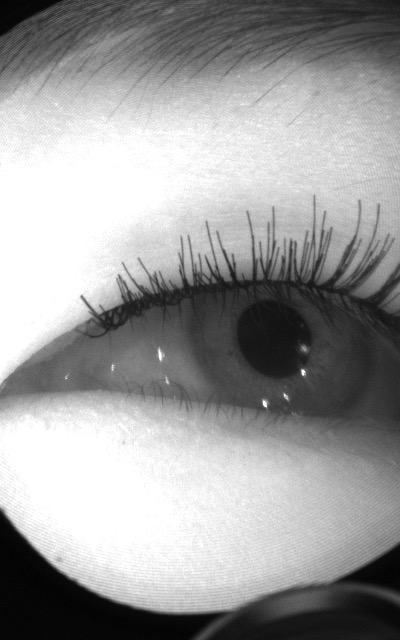} \hskip -0.3ex
    \includegraphics[width=4cm]{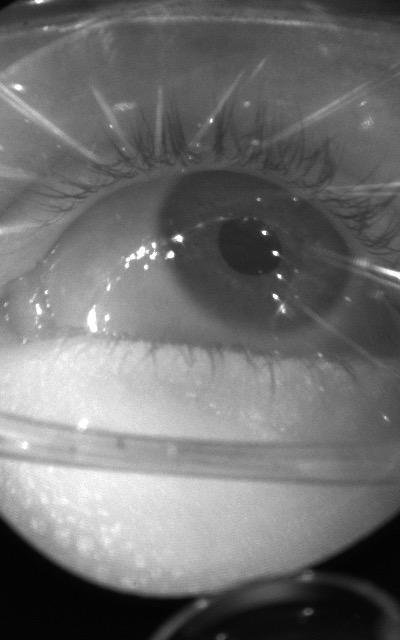} 
    \includegraphics[width=1.97cm]{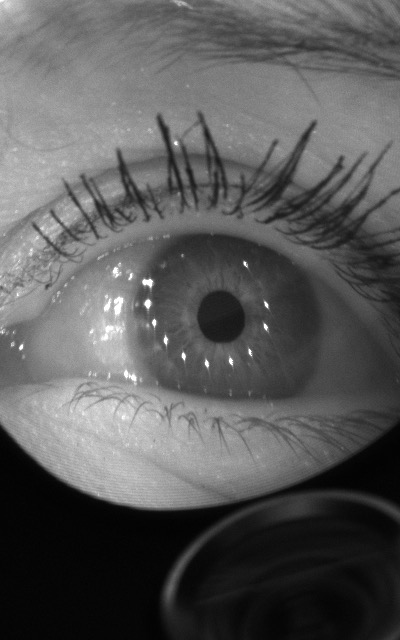} \hskip -0.3ex
    \includegraphics[width=1.97cm]{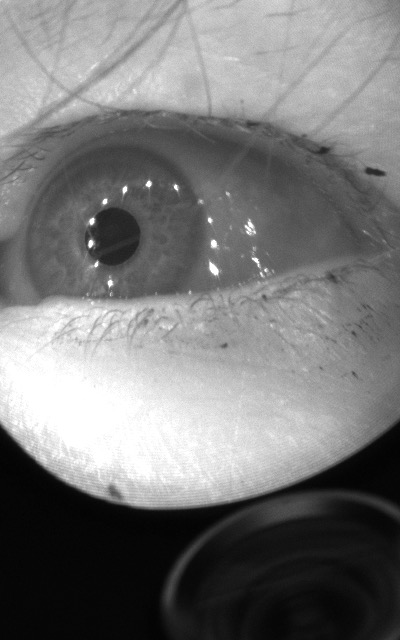} \hskip -0.3ex
    \includegraphics[width=1.97cm]{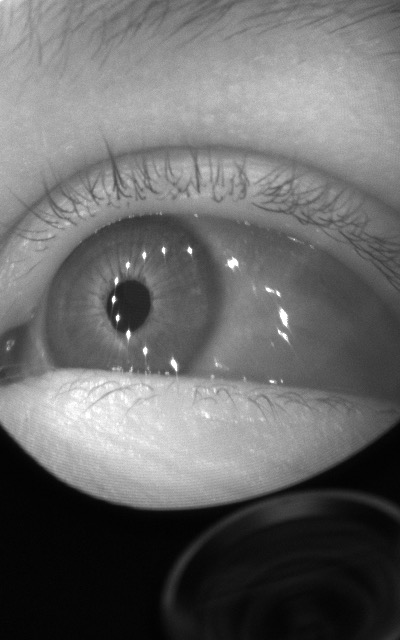} \hskip -0.3ex
    \includegraphics[width=1.97cm]{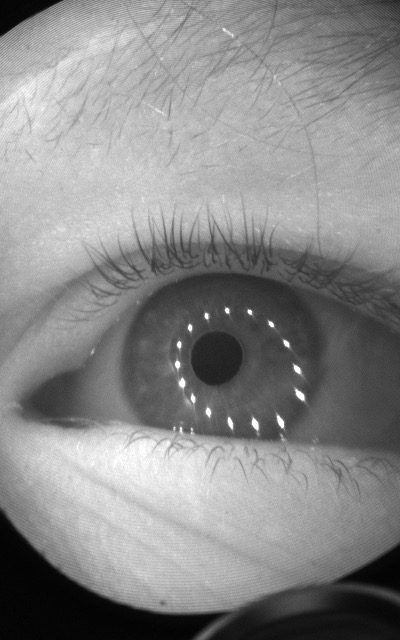} 
    \caption{Examples of the acquired HMD images.}
    \label{fig:hmd_image_examples}
\end{figure}


\begin{figure}
\centering
\subfigure{\includegraphics[width=2.7cm]{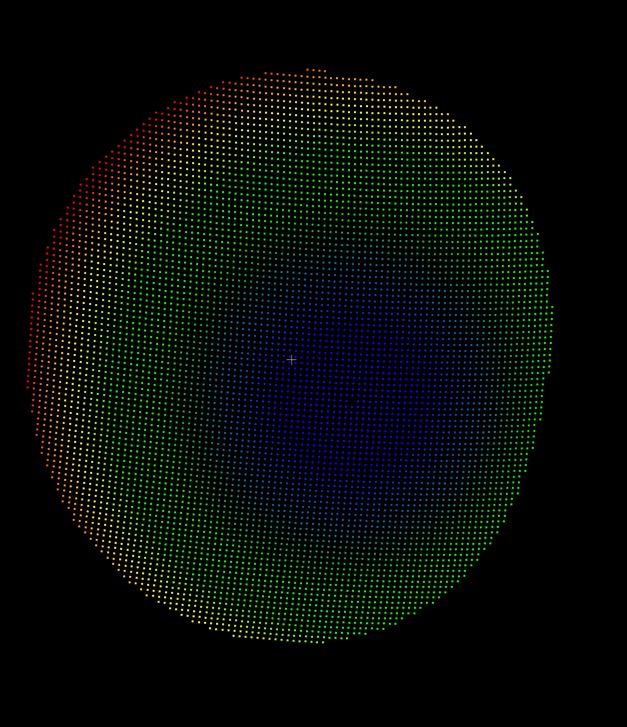}} \hskip -0.5ex
\subfigure{\includegraphics[width=2.7cm]{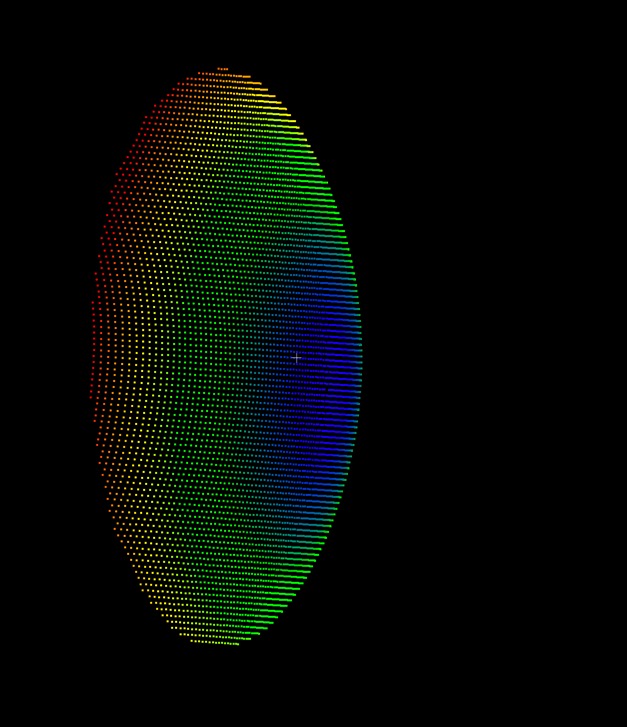}} \hskip -0.5ex
\subfigure{\includegraphics[width=2.7cm]{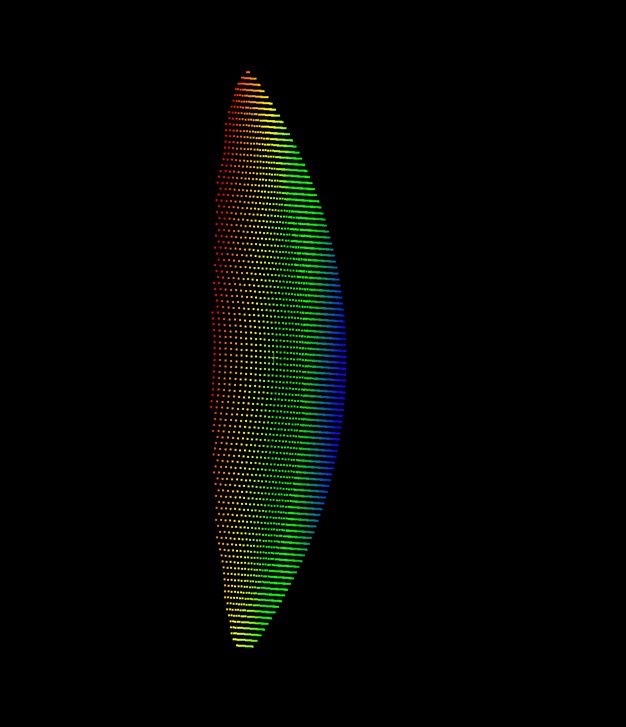}} \\
\vspace{-1\baselineskip}
\caption{Corneal topography of left eye.}
\vspace{-1.05em}
\subfigure{\includegraphics[width=2.7cm]{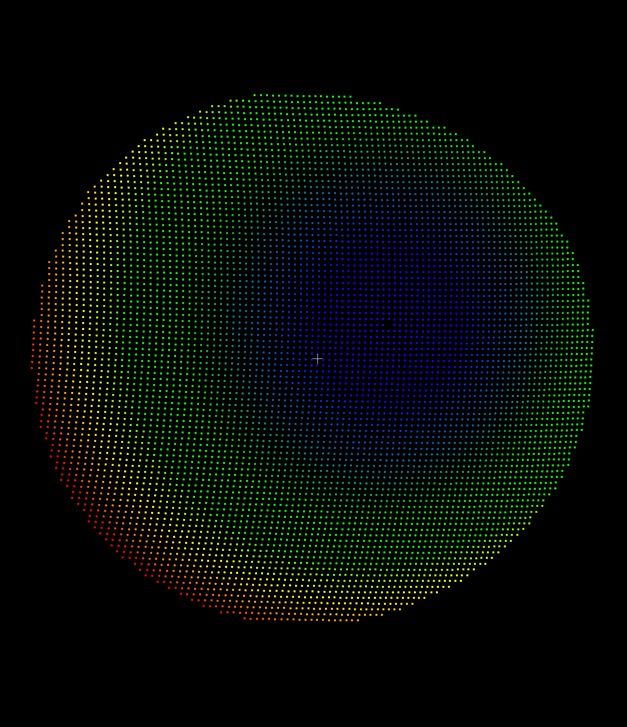}} \hskip -0.5ex
\subfigure{\includegraphics[width=2.7cm]{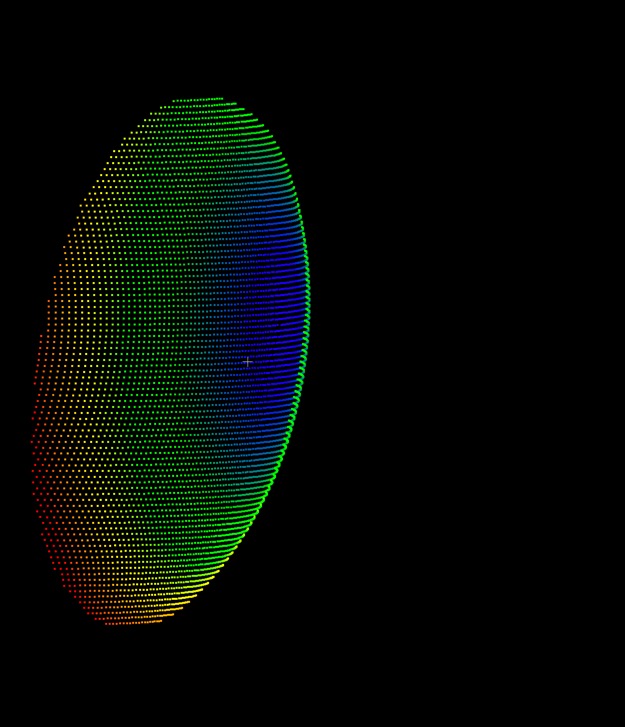}} \hskip -0.5ex
\subfigure{\includegraphics[width=2.7cm]{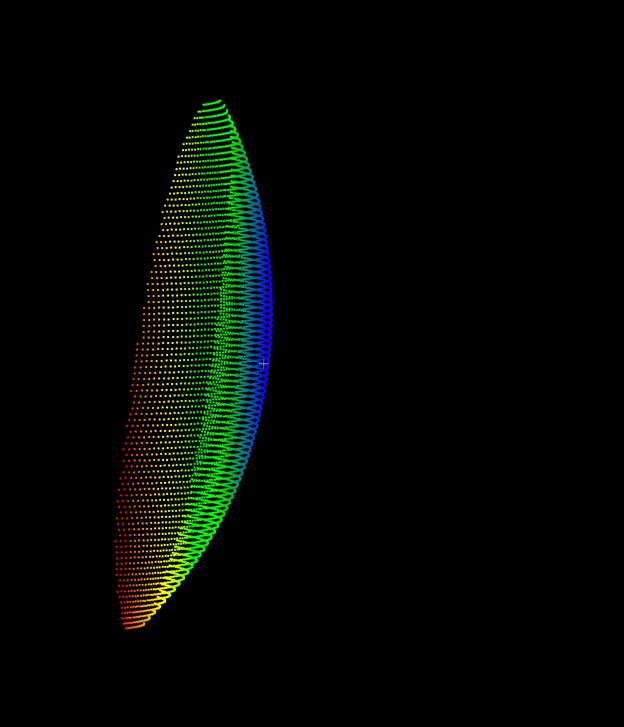}}
\caption{Examples of corneal topography, represented as point clouds. Row from top to bottom: left eye, right eye. Column from left to right: rotations along Y-axis. Color variation is along Z-axis. Better viewed in color.}
\label{fig:CT_examples}
\end{figure}

\section{Related Work}
\noindent
\textbf{Eye Tracking Datasets:}

\begin{table*}
    \centering
    \begin{tabular}{ |p{2.9cm}|p{1.5cm}|p{1.65cm}|p{1.5cm}|p{1.5cm}|p{1.5cm}|p{1.5cm}|p{1.5cm}|  }
        \hline
        Dataset & \#Images & \#Participants & Resolution & FrameRate & Controlled Light & Sync. Left/Right & Optometric Data\\ \hline
        STARE \cite{stareDatasetPaper} & - & - & - & - & No & No & Yes \\ \hline
        PoG \cite{mcmurrough2012eye} & - & 20 & - & 30 & Yes & No & No\\ \hline
        MASD \cite{previousCompetitionDataset} & 2,624 & 82 & - & - & No & No & No\\ \hline
        Ubiris v2 \cite{proenca2010ubiris} & 11,102 & 261 & 400$\times$300 & - & No & - & No\\ \hline
        LPW \cite{tonsen2016labelled} & 130,856 & 22 & 480$\times$640 & 95Hz & No & No & No\\ \hline
        NVGaze \cite{kim2019} & 2,500,000 & 30 & 480$\times$640 & 30Hz & Yes & No & No\\ \hline
        Gaze Capture \cite{cvpr2016_gazecapture} & 2,500,000 & 1,450 & Various & - & - & No & No\\ \hline
        Ours & 356,649 & 152 & 400$\times$640 & 200Hz & Yes & Yes & Yes \\ \hline
    \end{tabular}
    \caption{Publicly available datasets in the field of eye tracking. Note that studies in \cite{cvpr2016_gazecapture} and \cite{proenca2010ubiris} offer a large number of participants through crowd-sourcing, where images usually include a large portion of the face and hence lack eye details. }
    \label{tab:datasets_comparison}
\end{table*}

Due to the difficulty of capturing binocular eye data especially in the VR context, there exists only a limited number of large-scale high resolution human image datasets in this domain. A comparison of our dataset to some existing datasets can be found in Table~\ref{tab:datasets_comparison}.

The most similar dataset in terms of domain and image specifications is the recently published NVGaze dataset \cite{kim2019}, consisting of 2.5 million infrared images recorded from 30 participants using an HMD (640x480 at 30 Hz). NVGaze includes annotation masks for key eye-regions for an additional dataset of 2 million synthetic eye images but does not provide segmentation annotations for the human eye image set at this point. 
The LPW dataset \cite{tonsen2016labelled} includes a number of images recorded from 22 participants wearing a head mounted camera. 
Images are from indoor and outdoor recordings with varying lighting conditions and thus very different from the controlled lighting conditions in a VR HMD.

Some eye focused image sets are aimed at gaze prediction and released with gaze direction information, but do not include annotation masks, such as the Point of Gaze (PoG) dataset  \cite{mcmurrough2012eye}.
Other large-scale eye image datasets were captured  in the context of appearance-based gaze estimation and record the entire face using RGB cameras as opposed to the eye region \cite{funes2014eyediap, DBLP:journals/corr/HuangVS15, 7299081}. 
For example, Gaze Capture \cite{cvpr2016_gazecapture} consists of over 2.5 million images at various resolutions, recorded through crowd-sourcing on mobile devices, and its images are not specifically focused on the eye but contain a large portion of the surrounding face. 
In all these datasets, the focus is not solely on captures of eye-images, making them less suitable for the specific computer vision and machine learning challenges in the VR context. 
Finally, a different category of dataset, such as the UBIRIS \cite{proenca2010ubiris} and UBIRIS v2 \cite{ubiris2}, were conceived with iris recognition in mind and therefore contain only limited annotation mask information. \\

\noindent
\textbf{Eye Segmentation:} \\
Segmentation of ocular biometric traits in eye region, such as the pupil, iris, sclera, can provide the information to study fine grained details of eye movement such as saccade, fixation and gaze estimation \cite{venkateswarlu2003eye}.
A large amount of studies have been investigated on segmenting a single trait (e.g., only the iris, sclera or eye region) \cite{sankowski2010reliable, radu2015robust, thoma2016survey, das2017sserbc, lucio2018fully}.
A detailed survey of iris and sclera segmentation is presented in \cite{adegoke2013iris, das2013sclera}. We note that, although there are several advantages to having segmentation information on all key eye-regions simultaneously,  a very limited amount of studies have been done on multi-class eye segmentation \cite{rot2018deep,luo_et_al:OASIcs:2019:10188}.  However, study in \cite{rot2018deep} trained a convolutional encoder-decoder neural network on a small data set of 120 images from 30 participants. Study in \cite{luo_et_al:OASIcs:2019:10188} trained a convolutional neural network coupled with conditional random field for post-processing, on a data set of 3,161 low resolution images to segment only two classes: iris and sclera.
In this paper, we try to address the gap for multi-class eye segmentation including pupil, iris, sclera and background, in a large data set of images in high resolution of 400$\times$640.\\
\noindent
\textbf{Eye Rendering:} \\
\noindent
Generating eye appearances under various environmental conditions including facial expressions, color of the skin and the illumination settings,  play an important role in gaze estimation and tracking \cite{kim2019}. Two approaches have been studied in eye rendering: graphics-based approaches to generate eye images using a 3D eye model usually with a rendering framework to provide geometric variations such as gaze direction or head orientation \cite{wood2015rendering, wood2016learning}.
Another is machine learning based approach. 
Study in \cite{shrivastava2017learning} used a generative adversarial network to train models with synthetic eye images while testing on realistic images.
Study in \cite{wang2018hierarchical} used a conditional bidirectional generative adversarial network to synthesize eye images consistent with the given eye gaze.
However, these studies are focused on rendering synthetic eye images. We are interested in rendering realistic eye images that conform to the captured distribution of an individual and anticipate OpenEDS will encourage researchers to use the large corpus of eye-images and the annotation masks to develop solutions that can render realistic eye images.

\section{Data Collection}
To ensure eye-safety during our data-collection process, which is important because OpenEDS exposes users' eyes to infrared light, the exposure levels were controlled to be well below the maximum permissible exposure as laid out in IEC 62471:2006, as well as the American National Standard for the Safe use of Lasers (ANSI Z136.1-2000, (IEC) 60825).

OpenEDS was collected from voluntary participants of ages between $19$ and $65$. These participants have provided written informed consent for releasing their eye images before taking part in the study. There was no selection bias in our selection of participants for OpenEDS study, except that we required subjects to have corrected visual acuity above legal blindness, have a working knowledge of the English language, and not be pregnant. Participants were paid for their participation per session, and were given the choice to withdraw from the study at any point.

In addition to the image data captured from the HMD, OpenEDS also comes with an anonymized set of metadata data per participant:
\begin{itemize}
\item Age (19-65), sex (male/female), usage of glasses (yes/no);
\item Corneal topography.
\end{itemize}
Other than a pre-capture questionnaire for every participant, the rest of the data is captured in an approximately hour long session which is further split into sub-sessions as follows:  All optometric examination are conducted in the first 10  minutes, followed by a 5 minute break. The break is taken to prevent any effect of optometric examinations on data captured with the HMD. Finally, two 20 minute capture sessions using the HMD are used to capture the eye-data during which the users  were asked to perform several tasks such as focusing on specific points and other free-viewing experiments, separated by further breaks.  The images present in the initial release of the OpenEDS are taken from these 20 minute capture sessions.

\section{Annotation}
We generated annotation masks for key eye-regions from a total of 12,759 eye-images as follows: a) eyelid, (using human-annotated key points) b) Iris (using an ellipse), c) Iris (using human-annotated points on the boundary), d) Pupil (using an ellipse) and e) Pupil (using human-annotated key points). The key-point dataset was used to generate annotation masks, which are released as part of OpenEDS. Below, we provide more details into the annotation protocol that was followed in generating the key-points and the ellipse.

\begin{figure}
    \centering
    \includegraphics[width=8.5cm]{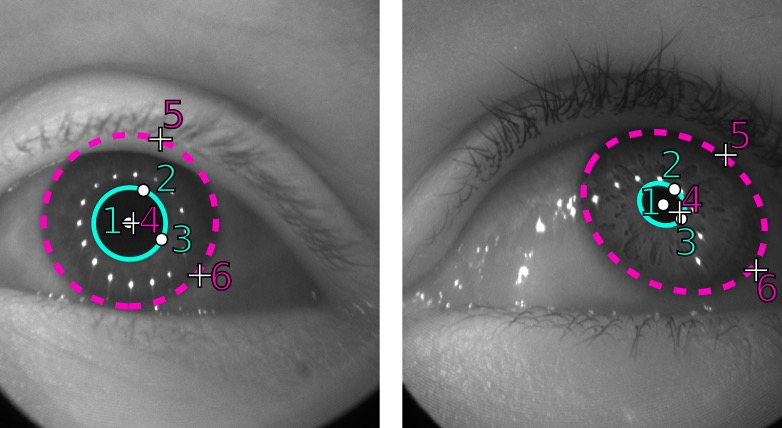}
    \caption{Ellipse Annotations. Points 1, 2 and 3 describe the Pupil, whereas points 4, 5 and 6 describe the Iris. Note that the points 1 and 4 are the center points. Best viewed digitally in color at high-resolution zoomed in.}
    \label{fig:ellipse_annotations}
\end{figure}

\subsection{Iris \& Pupil Annotation with Ellipses}
In order to produce the ellipse annotations for the Iris and the Pupil regions, the annotators performed the following three steps:
\begin{itemize}
\item Position the center point;
\item Position the second control point to adjust the shape and rotation; and
\item Position the third control point to constraint the remaining degree of freedom for ellipse fitting.
\end{itemize}
All points are colored and numbered appropriately in the annotation tool to avoid confusion. Figure \ref{fig:ellipse_annotations} offers an illustration of this process.

\begin{figure}
    \centering
    \includegraphics[width=8.5cm]{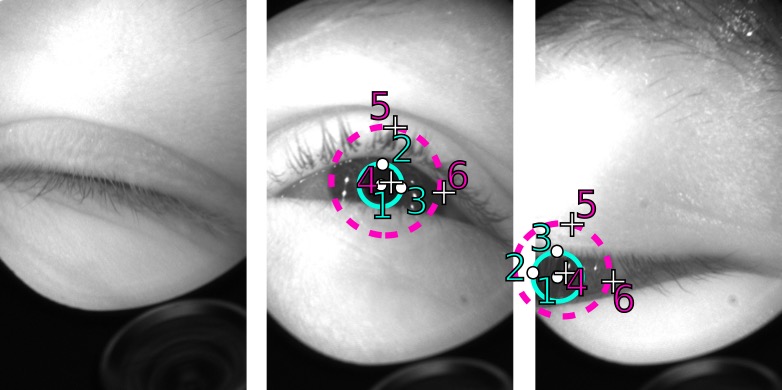}
    \caption{Ellipse annotations for difficult cases. For images where iris and pupil are not visible, no annotations are provided \textit{(left)}. However, if positions can be inferred, we obtain ellipses for each class \textit{(middle, right)} Please note that the images are padded for annotation so that ellipses can extend beyond the image boundaries \textit{(right)}. Best viewed digitally in color and zoomed in.}
    \label{fig:ellipse_annotations_skip}
\end{figure}

Since the ellipses do not provide any information about occlusion due to viewpoint and eyelid closure, we also obtain a more detailed annotation for which the annotators are asked to place a larger number control points, which allows us to extract complex polygonal regions with high accuracy. We instruct the annotators to skip the ellipse-based annotation if the iris and pupil are not visible, but allow ellipses if they can be inferred with reasonable certainty. Some of the difficult cases are shown in Figure \ref{fig:ellipse_annotations_skip}. Some of the images do not contain any useful information because the eyelids are completely closed or there is severe occlusion of the iris and pupil.  In these situations annotators are asked to skip labeling images. In particular, we do not annotate eye-images if:
\begin{itemize}
\item The eye is closed;
\item Eyelashes occlude the eye to the point where it is unreasonable to make an estimation;
\item The eye is out of view, which can happen e.g. if the HMD is misaligned in a particular frame. Typically, it happens when the user is either wearing it or adjusting it to their head.
\end{itemize}

\subsection{Iris \& Pupil Annotation with Polygons}
\begin{figure}
    \centering
    \includegraphics[width=8.5cm]{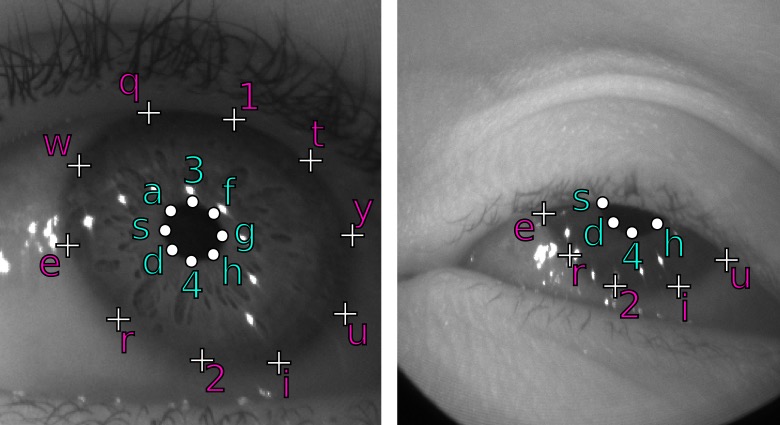}
    \caption{Iris and pupil annotations with dots (up to 10 points per feature). Top / bottom points are labelled with numbers, and further boundary points are denoted with letters. We note that even if one or the other of the top and bottom points is not visible, we still obtain annotations.}
    \label{fig:point_annotations}
\end{figure}

For the pupil and iris, we use 10 points each, to create a polygon annotation. In both cases, the process starts by placing two points at the top and bottom of the feature of interest (labelled with numbers) and spacing a further 4 points equally between them on the boundary (labelled with letters). The top and bottom points do not have to be placed, and, if both cannot be identified, we fall back to labelling just with the points for the left and right side. With this process, the annotators are instructed to proceed with the left side first, and then the right (see Figure \ref{fig:point_annotations}). Again, the annotators are asked to skip difficult-to-label images, as explained in Section 4.1.

\subsection{Eyelid Annotation}
\begin{figure}
    \centering
    \includegraphics[width=8.5cm]{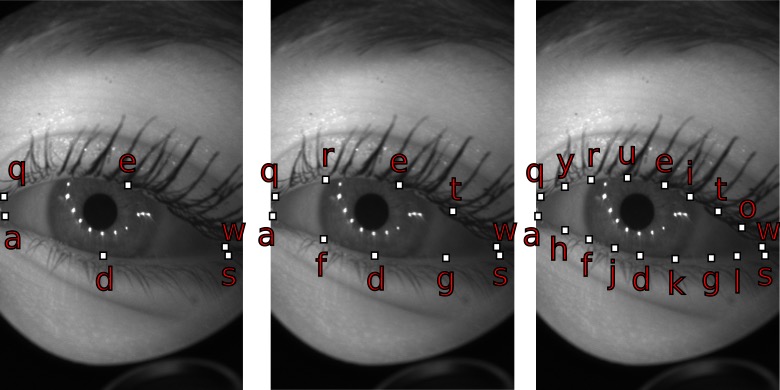}
    \caption{Eyelid annotation process. Note how the annotators proceed by splitting the annotation boundary recursively \textit{(left to right)}.}
    \label{fig:point_annotations_eyelid_procedure}
\end{figure}

\begin{figure}
    \centering
    \includegraphics[width=8.5cm]{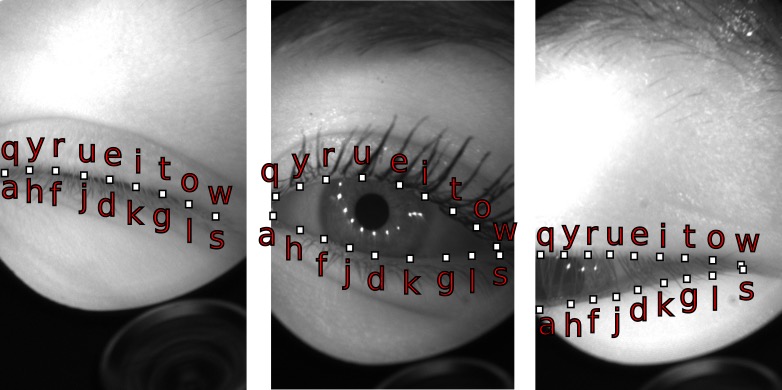}
    \caption{Eyelid annotation examples. Note how in case of the eye fully closed, we require the eyelid annotation points to overlap.}
    \label{fig:point_annotations_eyelid_correct}
\end{figure}

18 points are used for annotating the upper and lower eyelid. Similar to the iris and pupil case, the annotators are instructed to place these points equally spaced along the eyelid boundaries. Since the eyelid is much bigger than the other two cases (Iris and  Pupil), in order to help with equal spacing we give the instruction to split the line recursively while adding points. Examples of this process are given in Figure \ref{fig:point_annotations_eyelid_procedure}. Figure \ref{fig:point_annotations_eyelid_correct} shows completed annotations for a variety of cases.

\begin{table*}
  \begin{tabular}{| c | c |c |c |c |c |c |c |c |c || c | c | c || c | }
    \hline
    \multirow{2}{*}{} &
      \multicolumn{2}{c|}{Sex} & %
      \multicolumn{5}{c|}{Age} &%
      \multicolumn{2}{c||}{Glasses} & %
      \multicolumn{3}{c||}{\#Images} & %
      \multirow{2}{*}{\#CT.} \\
      \cline{2-13}
    & Female & Male & 18-23 & 24-30 & 31-40 & 41-50 & 51-65  & No & Yes & SeSeg. & IS. & Seq. & \\ \hline
    Train & 51& 44 & 3 & 15 & 39 & 26 & 12 & 92 & 3 & 8,916 & 193,882 & 57,000 &178 \\ \hline
    Val. & 13 & 15 & 1 & 5 & 5 & 8 & 9 & 27 & 1 & 2,403 & 57,337 & 16,800 & 52 \\ \hline
    Test & 18 & 11 & 4 & 3 & 7 & 6 & 9 & 27 & 2 & 1,440 & 1,471 & 17,400 & 56 \\ \hline
    Total & 82 & 70 & 8 & 23 & 51 & 40 & 30 & 146 & 6 & 12,759 & 252,690 & 91,200 & 286 \\ \hline
  \end{tabular}
  \caption{Statistics in OpenEDS for train, validation and test. SeSeg.: Images with semantic segmentation annotations. IS.: Images without annotations. Seq.: Image sequence set. CT: corneal topography. We report \#identities regarding demographics (sex, age), wearing glasses or not. We also report \#images and \#corneal topography (represented as point clouds).}
  \label{tab:stats_openeds}
\end{table*}

\section{General Statistics}
OpenEDS has a total of 12,759 images with annotation masks, 252,690 further images, 91,200 frames from contiguous 1.5 second video snippets, and 286 point cloud datasets for corneal topography. The latter image and image sequence sets are not accompanied by semantic segmentation annotations.
Table \ref{tab:stats_openeds} shows the statistics w.r.t. demographics (sex, age group, wearing glasses or not), and the amount of images and point clouds provided.

\subsection{Corneal Topography}
\label{subsec:ct}
We provide corneal topography point cloud data that captures the surface curvature of the cornea for both left and right eyes. Corneal topography of each participant was measured via Scheimpflug imaging using an OCULUS\textsuperscript{\textregistered} Pentacam\textsuperscript{\textregistered} HR corneal imaging system, where corneal elevation maps were exported and converted to a point cloud.
Figure \ref{fig:CT_examples} shows an example. We note that there is at most one point cloud estimate per participant.

\section{Statistical Analysis}\label{analysis}
As outlined above, we choose to split OpenEDS by identity of the study participants as we found this to be both intuitive, and  an easy setting to assess and avoid bias. When selecting the validation and test sets, we resample (or alternatively reweigh for evaluation) the data to account for factors such as age and sex. Resampling or weighting of this kind is motivated by the fact that under-sampled modes of the true data distribution are the hardest to accurately capture by data-driven approaches. To avoid bias arising from this, we ensure that the reweighing and selection of the test set penalizes approaches that do not take this into consideration.

\begin{figure}
    \centering
    \includegraphics[width=8cm]{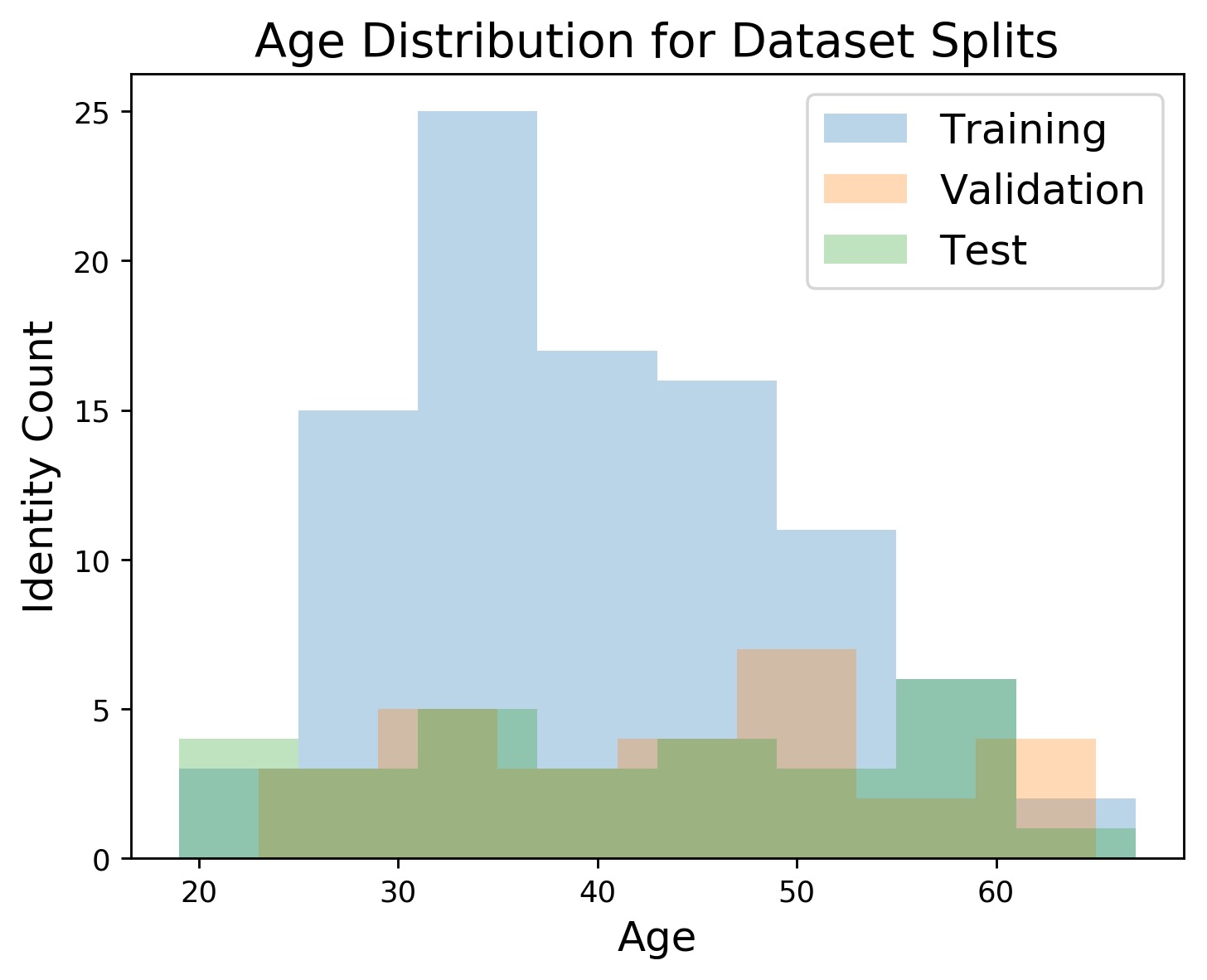}
    \caption{Age distribution of our proposed dataset splits. Note how we correct for bias in the collected data as far as possible.}
    \label{fig:age_whole_vs_test}
\end{figure}

An example of this is the age distribution. Figure \ref{fig:age_whole_vs_test} shows histograms characterizing the age distribution of the training vs test and validation images (with a bin-width of $5$ years). Note how our choice of dataset splits already removes a significant amount of bias.

Apart from the information we can directly gain from the collected metadata, we further investigate the dataset for forms of sampling bias. To do this, we take an imagenet \cite{imagenet_paper} pretrained 150-layer ResNet \cite{DBLP:journals/corr/HeZRS15}, as provided by the PyTorch library \cite{paszke2017automatic}, and encode a subset of 60,000 images sampled uniformly among the study participants to $4096$-dimensional feature space (the output of the second-to-last fully connected layer). We subsequently use the k-means implementation of FAISS \cite{DBLP:journals/corr/JohnsonDJ17} to cluster the encodings. Analysis of the resulting partitioning of the images reveals that most clusters predominantly contain images of either the left or right eye of \emph{one} identity. This supports the intuition that identities are distinctive, and that it is easy to tell apart the left and right side of the face. We are however more concerned with clusters that exhibit a similar proportion of left \emph{and} right eyes, or contain more than one identity. These clusters, if not corrected for in the process of sampling the identities used for each subset of the data splits, are evidence of a potentially significant difference of the distribution for the training, test and validation sets. The existence of such out of distribution cases could cast doubt over the validity of the test and validation process \cite{DBLP:journals/corr/LiangLS17}, and we therefore correct for it as far as possible.

Apart from identifying invalid images which can be trivially discarded (e.g. due to occlusion of the cameras or misalignment of the HMD), we identify the following difficult and under-sampled cases from the clustering process that are \emph{not} correlated with the previously identified factors of variation such as sex, age, and left/right differences:
\begin{enumerate}
\item Identities with glasses,
\item Images of almost closed eyes, and
\item Images of completely closed eyes.
\end{enumerate}
In order to make sure that identities with glasses are represented across all data splits, we provide additional per-user annotations of the glasses case and make sure that at least one such case is contained in the test and validation sets.

\begin{figure}
    \centering
    \includegraphics[width=2cm]{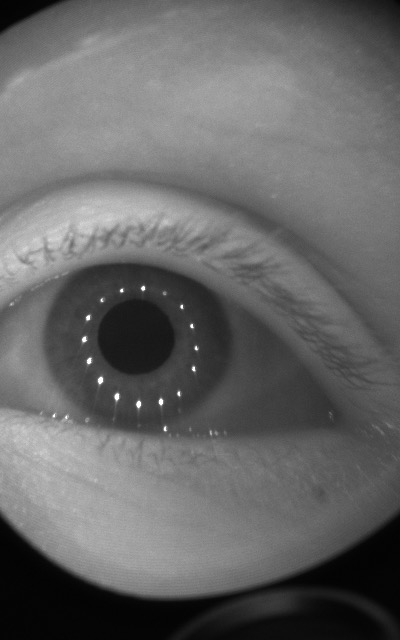}
    \includegraphics[width=2cm]{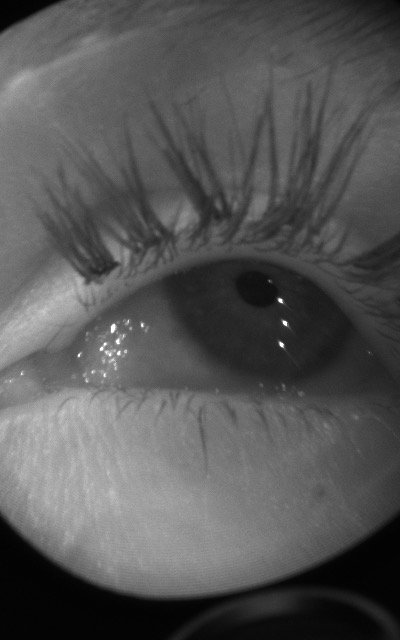}
    \includegraphics[width=2cm]{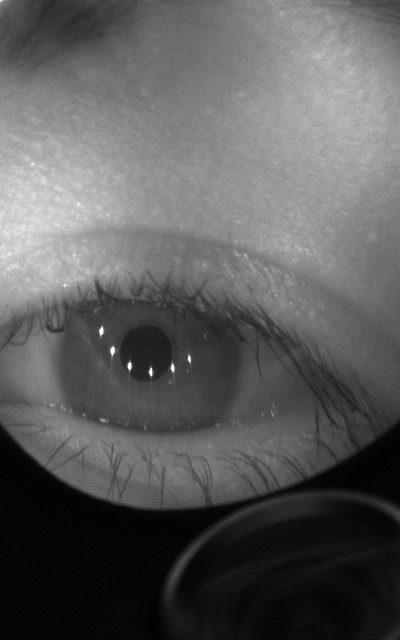}
    \includegraphics[width=2cm]{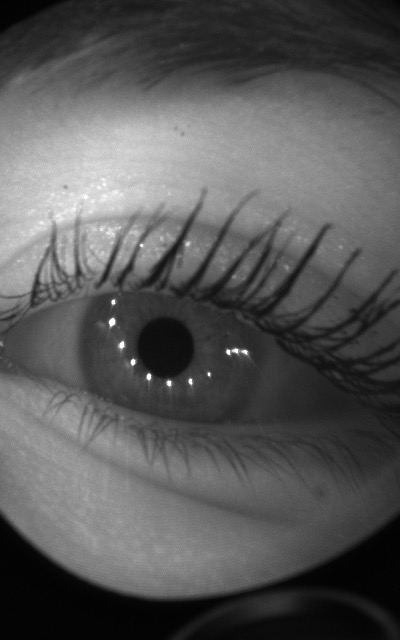}
    \includegraphics[width=2cm]{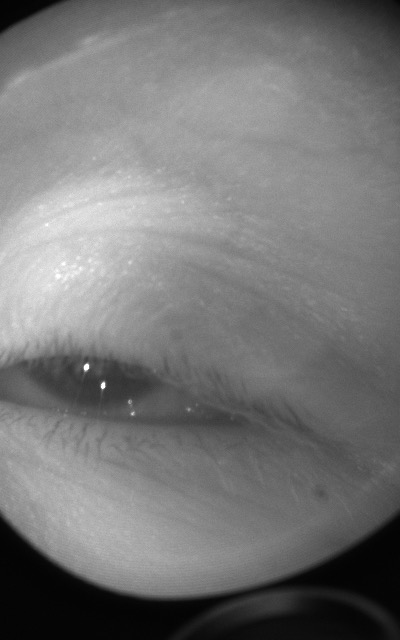}
    \includegraphics[width=2cm]{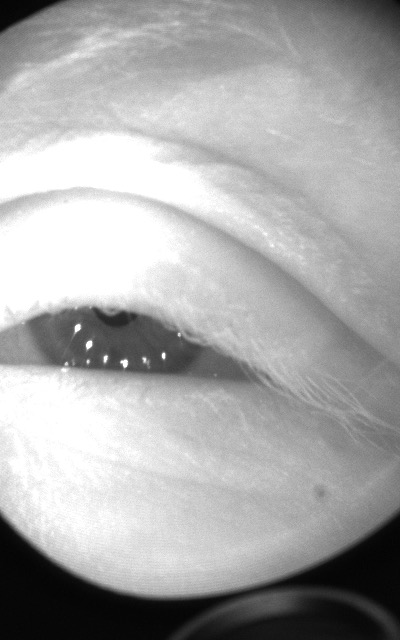}
    \includegraphics[width=2cm]{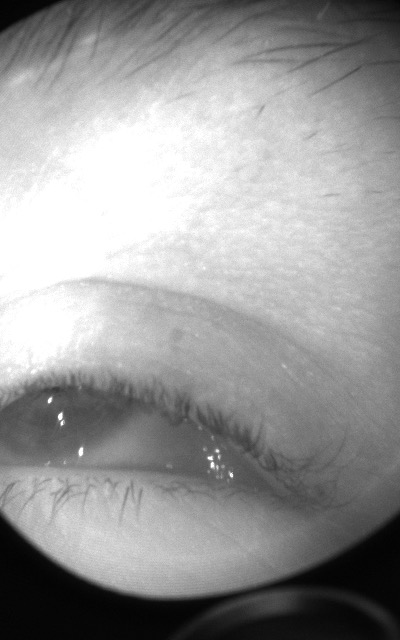}
    \includegraphics[width=2cm]{06222017EX0085_52589-left}
    \includegraphics[width=2cm]{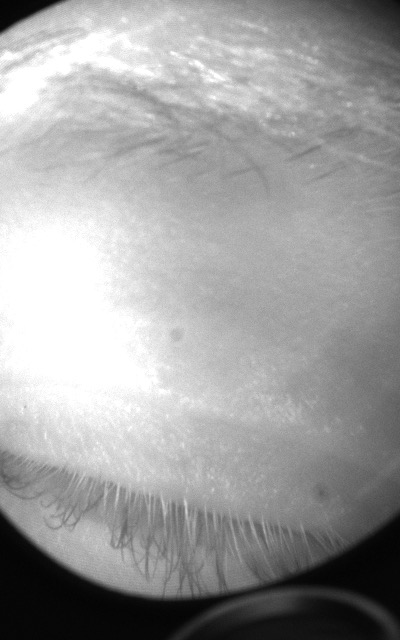}
    \includegraphics[width=2cm]{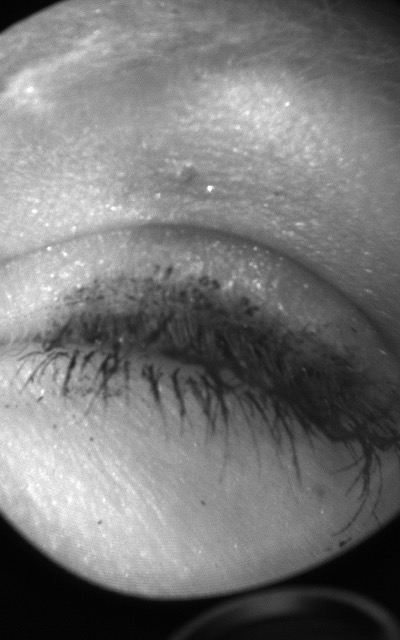}
    \includegraphics[width=2cm]{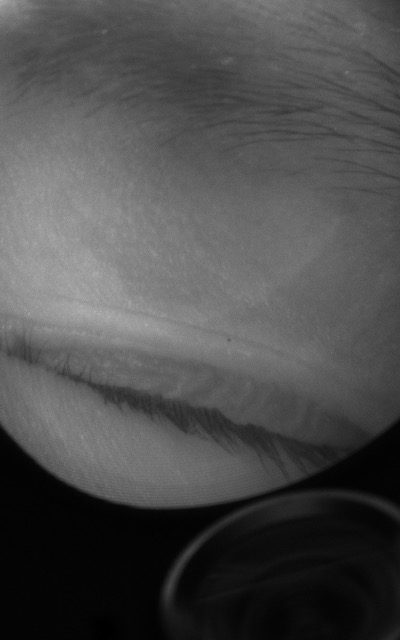}
    \includegraphics[width=2cm]{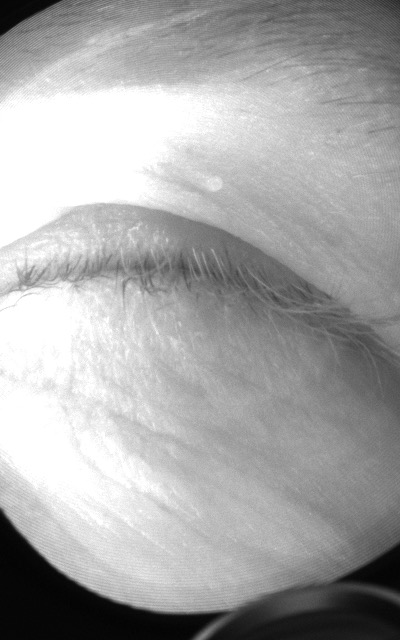}
    \includegraphics[width=2cm]{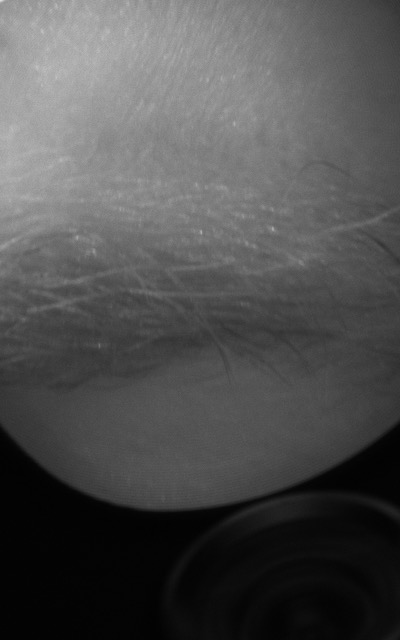}
    \includegraphics[width=2cm]{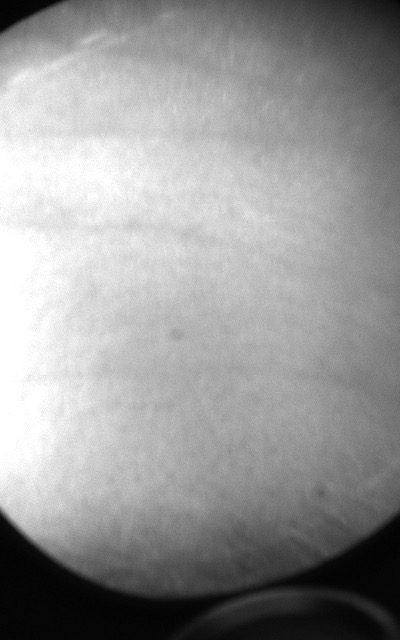}
    \includegraphics[width=2cm]{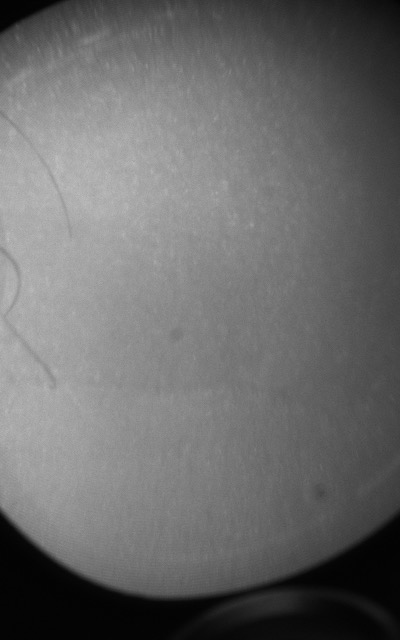}
    \includegraphics[width=2cm]{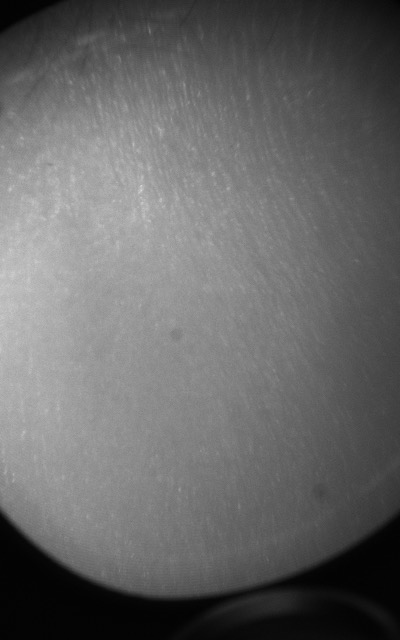}
    \caption{Representative images classified by our heuristic as (by row, from top to bottom) open, almost closed, closed and misaligned.}
    \label{fig:heuristic_examples}
\end{figure}

Ensuring that nearly and fully closed eyes are represented in OpenEDS is a more challenging problem. This is due to the volume of data and the fact that these cases cannot be easily delineated. For example, deciding what counts as \lq{nearly}\rq closed is not well defined. We address this by building a simple heuristic from a small ($<10000$) number of cases for \textit{open}, \textit{nearly closed}, and \textit{closed} eyes selected from the clusters described above. Example images given by this heuristic are shown in Figure~\ref{fig:heuristic_examples}.

We train a 50-layers ResNet to classify a subset of $2.5$ million images from the data, and use this to manually improve the small initial dataset. After repeating this process $4$ times, we achieve more than $~96\%$ accuracy on the heuristic. We find that this heuristic finds approximately twice as many \lq{nearly}\rq than fully closed eyes (on a subset of $12.6$ million images, we estimate the percentages for these cases to be $~2.21\%$ and $~4.31\%$, respectively). This is expected as the eye is near-fully closed just before and after blinking, thus providing evidence for the usefulness of our heuristic.

We make sure that all three identified eye states are present in the data when selecting images for OpenEDS but do not absolutely guarantee that any particular ratios are preserved.

\subsection{Identity-Centric Dataset Splits}
We partition OpenEDS in several ways that allow for the principled evaluation of a variety of machine learning problems, as shown in Table~\ref{tab:stats_openeds}. Reasoning from an identity-centric perspective, let $U$ be the set of all participants. We require semantic segmentation training, test and validation sets
\begin{equation}
    ss_{train} \subset U, ss_{val} \subset U, ss_{test} \subset U,
\end{equation}
where $ss_{train} \cap ss_{val} \cap ss_{test}= \emptyset$.


The set of users from which we uniformly select images for the additional image and sequence datasets \text{without} annotation, $E \subset U$, can contain images from users contained in $ss_{train}$, $ss_{test}$ and $ss_{val}$. The point of additionally providing $E$ is to encourage the user of unlabelled data to improve the performance of supervised learning approaches.

For the case of semantic segmentation, we roughly follow the ratio of popular datasets such as MSCOCO \cite{DBLP:journals/corr/LinMBHPRDZ14}. We thus have, from a user-centric perspective, a $ss_{train} / ss_{val} / ss_{test}$ split of $\frac{95}{152} / \frac{28}{152} / \frac{29}{152}$.



\begin{figure}[h]
    \centering
    \includegraphics[width=8cm]{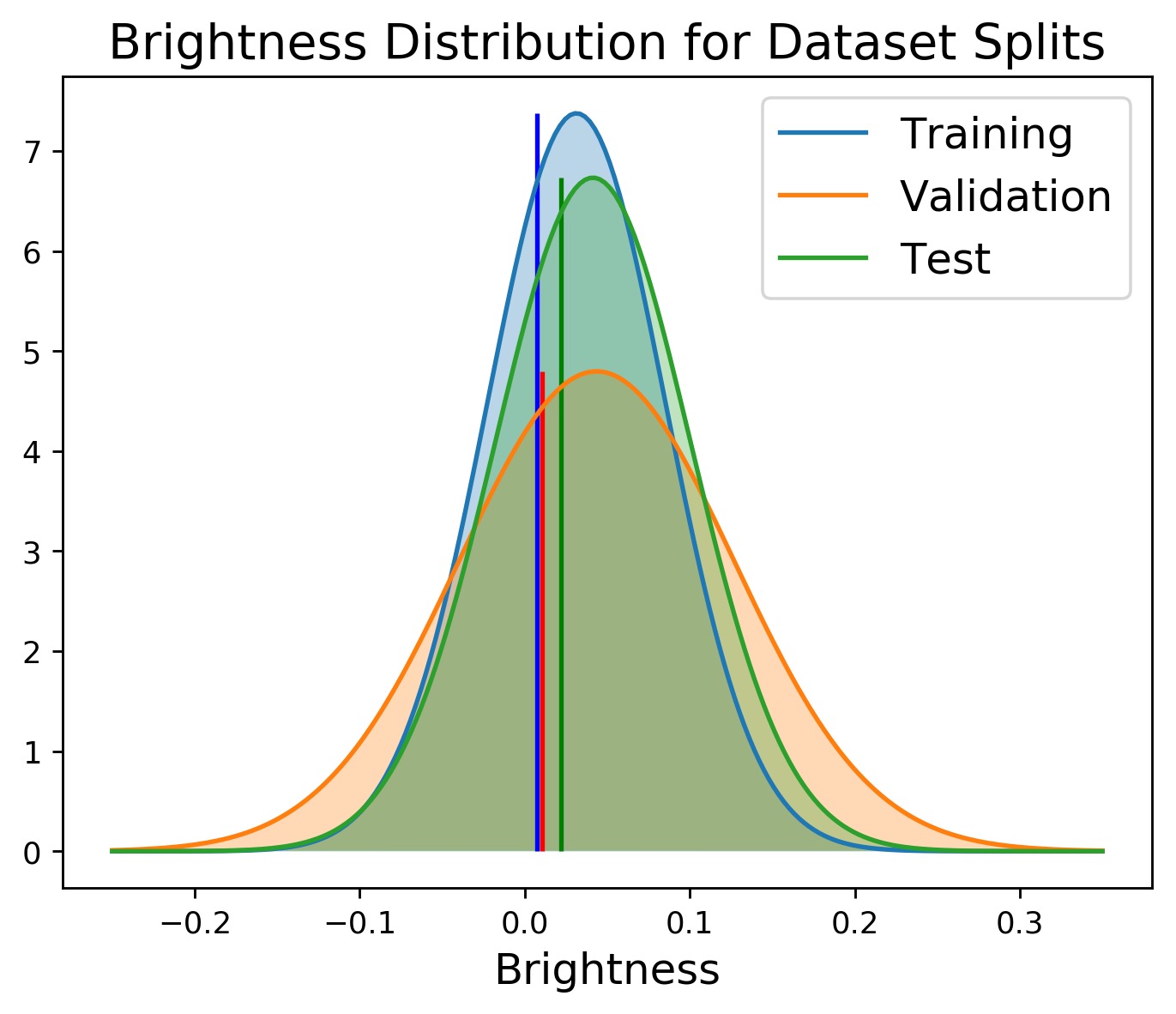}
    \caption{Per split distribution of the the mean image luminance. The median is denoted by vertical lines. Note that we do not apply any transformations to equalize these.}
    \label{fig:image_luminance_distribution}
\end{figure}

One characteristic of our image domain we leave intentionally unaltered for the OpenEDS release, is the image brightness distribution. As shown in Fig \ref{fig:image_luminance_distribution}, this can vary strongly on a per-identity basis. The reason for this are individual-specific reflectance properties of human skin and eye, the fit of the HMD (which can vary depending on the shape of an individual's face), and whether or not makeup is applied. We provide this information to encourage future analysis to focus on generalizing to brightness variations of this kind.


\subsection{Dataset Generation}
Generating the OpenEDS splits requires us to solve a constrained discrete optimization problem, subject to the constraints outlined in this section. We simplify this process by greedily selecting $ss_{test}, ss_{val}$ and $ss_{train}$, in that order. We obtain the individual balanced selections using 10 million random samples each, and picking the most balanced configuration among those samples. We found that this provides well-balanced splits. The results of this procedure can be seen in Figure \ref{fig:age_whole_vs_test}, where the age distribution is significantly more well-balanced in the test and validation as opposed to the training set.

As a final sanity check, we compute the Maximum Mean Discrepancy (MMD) (\cite{sriperumbudur2010hilbert}) between the training, validation and test sets to ensure the identity-based splits of the data produce sufficiently correlated subsets. The standard deviation of the Gaussian RBF kernel ($k(x, x\sp{\prime}) = \exp(-\frac{\lVert x - x\sp{\prime}\rVert^{2}_{2}}{2\sigma^{2}})$) used in computing the MMD is set based on the median Euclidian norm between all image pairs in the dataset \cite{DBLP:journals/corr/LiangLS17}. In our case, this amounts to setting $\frac{1}{2\sigma^2} =  0.0087719$. Evaluating the metric for random sets of 1024 images across the data splits gives the results shown in Table 3, which strongly indicates that the three derived image sets are drawn from the same underlying data distribution.

\begin{table}[]
\begin{tabular}{|l|l|l|l|}
\hline
           & Train    & Validation & Test    \\ \hline
Train      & 0.0      & 0.003932   & 0.00487 \\ \hline
Validation & 0.003932 & 0.0        & 0.00489 \\ \hline
Test       & 0.00487  & 0.00489    & 0.0     \\ \hline
\end{tabular}
\caption{Maximum Mean Disrepancy between the training, test and validation sets (Evaluated on a random set of 1024 images per split, averaged over 10 runs)}
\end{table}

\section{Investigation into neural network model to predict key eye-region annotation masks}
\begin{figure}
\centering
\subfigure{\includegraphics[width=2cm]{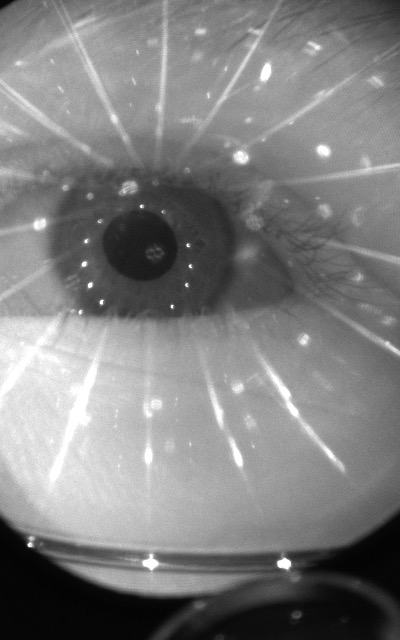}}
\subfigure{\includegraphics[width=2cm]{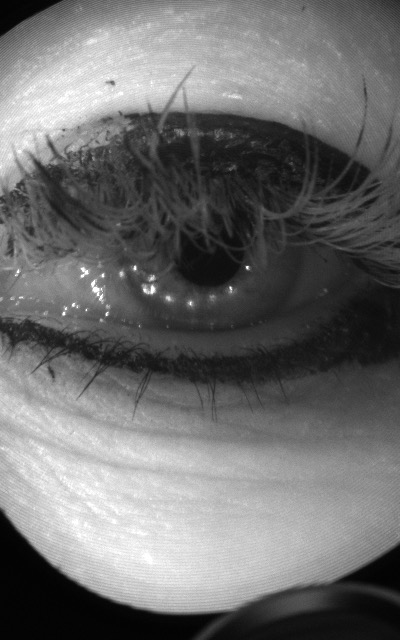}}
\subfigure{\includegraphics[width=2cm]{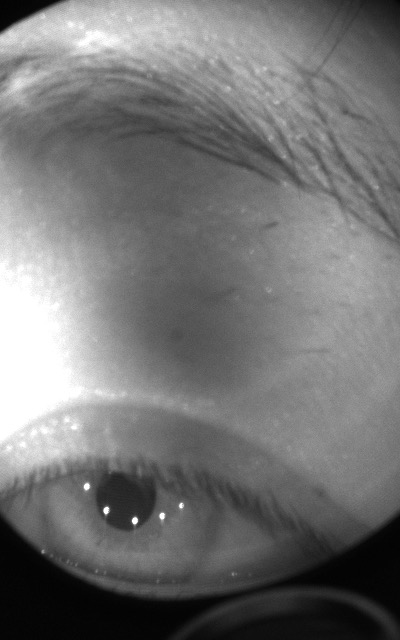}}
\subfigure{\includegraphics[width=2cm]{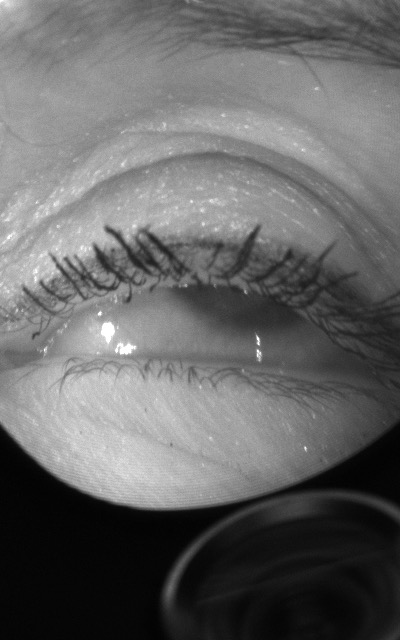}}\\
\vspace{-0.8em}
\subfigure{\includegraphics[width=2cm]{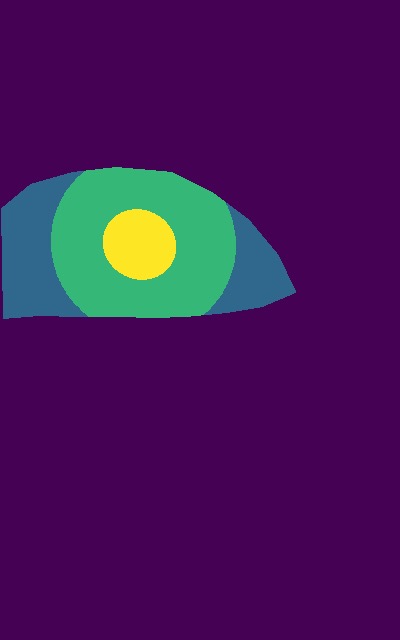}}
\subfigure{\includegraphics[width=2cm]{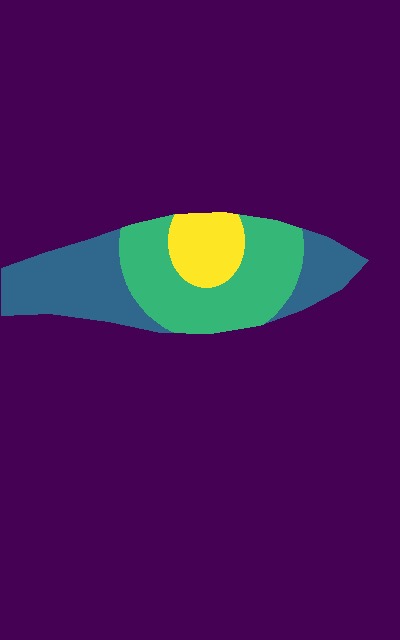}}
\subfigure{\includegraphics[width=2cm]{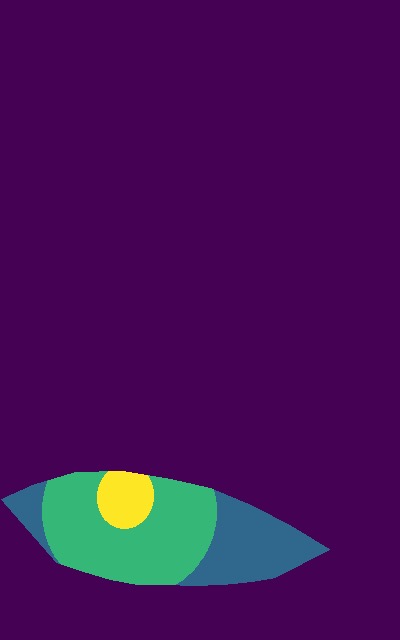}}
\subfigure{\includegraphics[width=2cm]{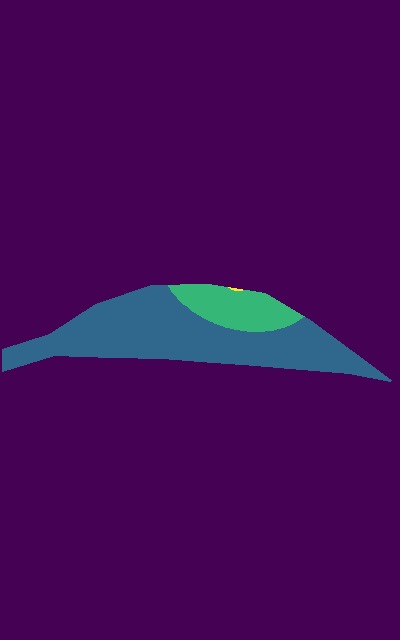}}\\
\vspace{-0.8em}
\subfigure{\includegraphics[width=2cm]{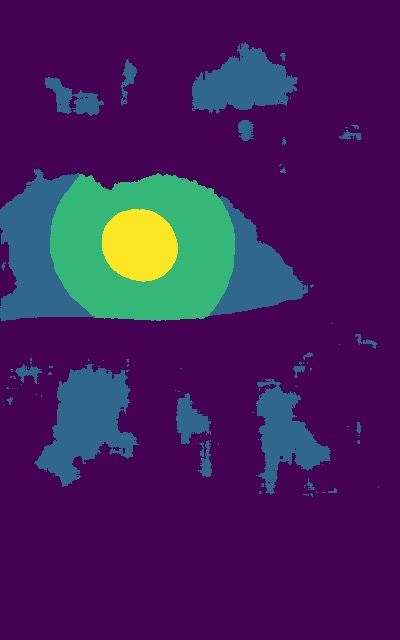}}
\subfigure{\includegraphics[width=2cm]{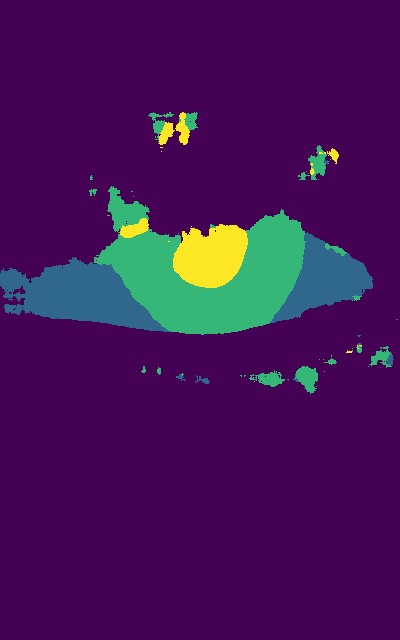}}
\subfigure{\includegraphics[width=2cm]{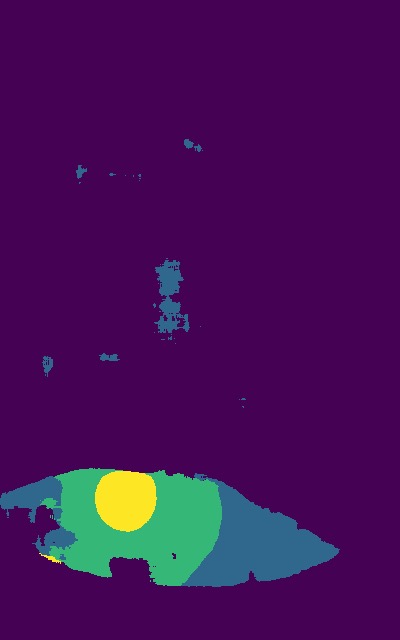}}
\subfigure{\includegraphics[width=2cm]{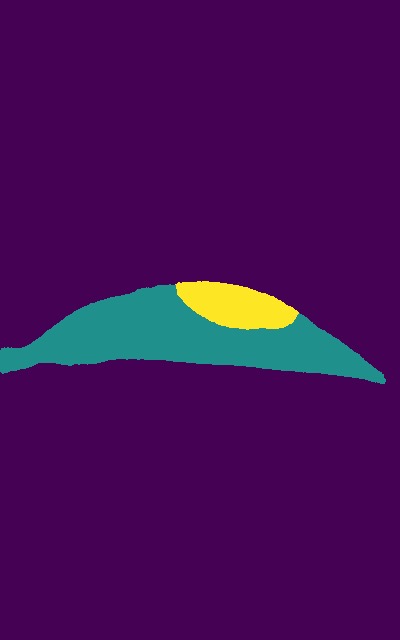}}
\caption{Examples of challenging samples on test data set of semantic segmentation. Rows from top to bottom: images, ground truth, predictions from SegNet w/ BR. Columns from left to right: eyeglasses, heavy mascara, dim light, varying pupil size. Better viewed in color.}
\label{fig:SS_examples}
\end{figure}

A number of convolutional encoder-decoder neural network architectures, derived from the SegNet architecture, \cite{badrinarayanan2017segnet}, (keeping top 4 layers from SegNet in the encoder and the decoder network) are investigated for predicting key eye-segments: boundary refinement (BR) modeling the boundary alignment as a residual structure to improve localization performance near object boundaries \cite{peng2017large}; separable convolution (SC) factorizing a standard convolution into a depth wise convolution and a 1$\times$1 convolution to reduce computational cost \cite{howard2017mobilenets}.  In addition, we introduced a multiplicative skip connection between the last layer of the encoder and the decoder network. We refer to this modified Segnet neural network as the mSegnet model \\
\indent We trained each network for 200 epochs on a NVIDIA RTX 2080 GPU using PyTorch \cite{paszke2017automatic} with ADAM optimizer, initial learning rate 0.0001, batch size 8, and weight decay $1\mathrm{e}{-5}$.
No data augmentation was performed.  \\
\begin{table}[t]
    \centering
    \begin{tabular}{|p{1.1cm}|p{0.7cm}|p{0.7cm}|p{0.7cm}|p{0.7cm}|p{0.8cm}|p{0.95cm}|}
        \hline
        Model & Pixel & Mean & Mean & Mean & Model  & \#Param. \\ 
        		  & acc. & acc. & F1 &  IoU &  Size (MB) & (M) \\ \hline
        mSegnet  & 98.0  & 96.8 & 97.9 & 90.7 & 13.3 & 3.5 \\
        \hline
        mSegnet w/ BR & 98.3  & 97.5 & 98.3 & 91.4 & 13.3 & 3.5 \\
        \hline
        mSegnet w/ SC & 97.6  & 96.6 & 97.4 & 89.5 & 1.6 & 0.4 \\
        \hline
    \end{tabular}
    \caption{Results on semantic segmentation. Model size: size of the PyTorch models on disk. \#Params: the number of learnable parameters, where “M” stands for million. {\it mSegnet}: 4-layer segnet with multiplicative skip connection between the last layer of the encoder and the decoder network}
    \label{tab:SS_accuracy}
\end{table}
\indent Table \ref{tab:SS_accuracy} shows the results of semantic segmentation.
We follow the same evaluation metric as in \cite{long2015fully}.
In addition, $\#$parameters are considered as well.
In terms of accuracy, SegNet with BR achieved the best performance with a mean IoU of 91.4\%.
In terms of model complexity, as measured in terms of the model size, SegNet with SC requires the smallest model size of 1.6 MB and the number of trainable parameters of 0.4 M. We also observed that in general all these models  fail to generate high fidelity eye-region masks for eye-images with eyeglasses and heavy mascara and non-conforming pupil orientation. Figure \ref{fig:SS_examples} shows examples from these failure cases.

\section{Conclusion}
We have presented OpenEDS, a new dataset of images and optometric data of the human eye. Initial algorithmic analysis of the provided labels demonstrates the usefulness of the data for semantic segmentation and we hope that future work can improve on these results. Our statistical analysis of the data and the resulting dataset splits should be of use to density estimation tasks for the eye-tracking domain.

\section*{Acknowledgement}
The authors gratefully acknowledge the helpful comments and feedback from Abhishek Sharma, Alexander Fix, and Nick Sharp.


\newpage
{\small
\bibliographystyle{ieee}
\bibliography{egbib}

\begin{thebibliography}{10}\itemsep=-1pt

\bibitem{adegoke2013iris}
B.~Adegoke, E.~Omidiora, S.~Falohun, and J.~Ojo.
\newblock Iris segmentation: a survey.
\newblock {\em International Journal of Modern Engineering Research (IJMER)},
  3(4):1885--1889, 2013.

\bibitem{badrinarayanan2017segnet}
V.~Badrinarayanan, A.~Kendall, and R.~Cipolla.
\newblock Segnet: A deep convolutional encoder-decoder architecture for image
  segmentation.
\newblock {\em IEEE transactions on pattern analysis and machine intelligence},
  39(12):2481--2495, 2017.

\bibitem{sriperumbudur2010hilbert}
K.~Borgwardt, A.~Gretton, M.~Rasch, H.~P. Kriegel, B.~Scholkopf, and A.~Smola.
\newblock Integrating structured biological data by kernel maximum mean
  discrepency.
\newblock {\em Bioinformatics}, 22:e49--457, 2010.

\bibitem{borji2013state}
A.~Borji and L.~Itti.
\newblock State-of-the-art in visual attention modeling.
\newblock {\em IEEE transactions on pattern analysis and machine intelligence},
  35(1):185--207, 2013.

\bibitem{das2013sclera}
A.~Das, U.~Pal, M.~Blumenstein, and M.~A.~F. Ballester.
\newblock Sclera recognition-a survey.
\newblock In {\em 2013 2nd IAPR Asian Conference on Pattern Recognition}, pages
  917--921. IEEE, 2013.

\bibitem{das2017sserbc}
A.~Das, U.~Pal, M.~A. Ferrer, M.~Blumenstein, D.~{\v{S}}tepec, P.~Rot,
  {\v{Z}}.~Emer{\v{s}}i{\v{c}}, P.~Peer, V.~{\v{S}}truc, S.~A. Kumar, et~al.
\newblock Sserbc 2017: Sclera segmentation and eye recognition benchmarking
  competition.
\newblock In {\em 2017 IEEE International Joint Conference on Biometrics
  (IJCB)}, pages 742--747. IEEE, 2017.

\bibitem{previousCompetitionDataset}
A.~{Das}, U.~{Pal}, M.~A. {Ferrer}, M.~{Blumenstein}, D.~{{\v S}tepec},
  P.~{Rot}, Z.~{Emersic}, P.~{Peer}, V.~{{\v S}truc}, S.~V.~A. {Kumar}, and
  B.~S. {Harish}.
\newblock Sserbc 2017: Sclera segmentation and eye recognition benchmarking
  competition.
\newblock In {\em 2017 IEEE International Joint Conference on Biometrics
  (IJCB)}, pages 742--747, Oct 2017.

\bibitem{imagenet_paper}
J.~Deng, W.~Dong, R.~Socher, L.~Li, K.~Li, and L.~Fei-Fei.
\newblock Imagenet: A large-scale hierarchical image database.
\newblock In {\em 2009 IEEE Conference on Computer Vision and Pattern
  Recognition}, pages 248--255, June 2009.

\bibitem{funes2014eyediap}
K.~A. Funes~Mora, F.~Monay, and J.-M. Odobez.
\newblock Eyediap: A database for the development and evaluation of gaze
  estimation algorithms from rgb and rgb-d cameras.
\newblock In {\em Proceedings of the Symposium on Eye Tracking Research and
  Applications}, pages 255--258. ACM, 2014.

\bibitem{geiger2013vision}
A.~Geiger, P.~Lenz, C.~Stiller, and R.~Urtasun.
\newblock Vision meets robotics: The kitti dataset.
\newblock {\em The International Journal of Robotics Research},
  32(11):1231--1237, 2013.

\bibitem{geiger2012we}
A.~Geiger, P.~Lenz, and R.~Urtasun.
\newblock Are we ready for autonomous driving? the kitti vision benchmark
  suite.
\newblock In {\em 2012 IEEE Conference on Computer Vision and Pattern
  Recognition}, pages 3354--3361. IEEE, 2012.

\bibitem{harezlak2018application}
K.~Harezlak and P.~Kasprowski.
\newblock Application of eye tracking in medicine: a survey, research issues
  and challenges.
\newblock {\em Computerized Medical Imaging and Graphics}, 65:176--190, 2018.

\bibitem{DBLP:journals/corr/HeZRS15}
K.~He, X.~Zhang, S.~Ren, and J.~Sun.
\newblock Deep residual learning for image recognition.
\newblock {\em CoRR}, abs/1512.03385, 2015.

\bibitem{EyeTrackingBook0}
K.~Holmqvist, M.~Nystr{\"o}m, R.~Andersson, R.~Dewhurst, H.~Jarodzka, and
  J.~van~de Weijer.
\newblock Eye tracking: A comprehensive guide to methods and measures.
\newblock 01 2011.

\bibitem{stareDatasetPaper}
A.~D. Hoover, V.~Kouznetsova, and M.~Goldbaum.
\newblock Locating blood vessels in retinal images by piecewise threshold
  probing of a matched filter response.
\newblock {\em IEEE Transactions on Medical Imaging}, 19(3):203--210, March
  2000.

\bibitem{howard2017mobilenets}
A.~G. Howard, M.~Zhu, B.~Chen, D.~Kalenichenko, W.~Wang, T.~Weyand,
  M.~Andreetto, and H.~Adam.
\newblock Mobilenets: Efficient convolutional neural networks for mobile vision
  applications.
\newblock {\em arXiv preprint arXiv:1704.04861}, 2017.

\bibitem{DBLP:journals/corr/HuangVS15}
Q.~Huang, A.~Veeraraghavan, and A.~Sabharwal.
\newblock Tabletgaze: {A} dataset and baseline algorithms for unconstrained
  appearance-based gaze estimation in mobile tablets.
\newblock {\em CoRR}, abs/1508.01244, 2015.

\bibitem{DBLP:journals/corr/JohnsonDJ17}
J.~Johnson, M.~Douze, and H.~J{\'{e}}gou.
\newblock Billion-scale similarity search with gpus.
\newblock {\em CoRR}, abs/1702.08734, 2017.

\bibitem{kim2019}
J.~Kim, M.~Stengel, A.~Majercik, S.~De~Mello, S.~Laine, M.~McGuire, and
  D.~Luebke.
\newblock Nvgaze: An anatomically-informed dataset for low-latency, near-eye
  gaze estimation.
\newblock In {\em Proceedings of the SIGCHI Conference on Human Factors in
  Computing Systems}, CHI '19, New York, NY, USA, 2019. ACM.

\bibitem{cvpr2016_gazecapture}
K.~Krafka, A.~Khosla, P.~Kellnhofer, H.~Kannan, S.~Bhandarkar, W.~Matusik, and
  A.~Torralba.
\newblock Eye tracking for everyone.
\newblock In {\em IEEE Conference on Computer Vision and Pattern Recognition
  (CVPR)}, 2016.

\bibitem{Kramida2016}
G.~Kramida.
\newblock Resolving the vergence-accommodation conflict in head-mounted
  displays.
\newblock {\em IEEE transactions on visualization and computer graphics},
  22(7):1912--1931, 2016.

\bibitem{krizhevsky2012imagenet}
A.~Krizhevsky, I.~Sutskever, and G.~E. Hinton.
\newblock Imagenet classification with deep convolutional neural networks.
\newblock In {\em Advances in neural information processing systems}, pages
  1097--1105, 2012.

\bibitem{DBLP:journals/corr/LiangLS17}
S.~Liang, Y.~Li, and R.~Srikant.
\newblock Principled detection of out-of-distribution examples in neural
  networks.
\newblock {\em CoRR}, abs/1706.02690, 2017.

\bibitem{DBLP:journals/corr/LinMBHPRDZ14}
T.~Lin, M.~Maire, S.~J. Belongie, L.~D. Bourdev, R.~B. Girshick, J.~Hays,
  P.~Perona, D.~Ramanan, P.~Doll{\'{a}}r, and C.~L. Zitnick.
\newblock Microsoft {COCO:} common objects in context.
\newblock {\em CoRR}, abs/1405.0312, 2014.

\bibitem{long2015fully}
J.~Long, E.~Shelhamer, and T.~Darrell.
\newblock Fully convolutional networks for semantic segmentation.
\newblock In {\em Proceedings of the IEEE conference on computer vision and
  pattern recognition}, pages 3431--3440, 2015.

\bibitem{lucio2018fully}
D.~R. Lucio, R.~Laroca, E.~Severo, A.~S. Britto~Jr, and D.~Menotti.
\newblock Fully convolutional networks and generative adversarial networks
  applied to sclera segmentation.
\newblock {\em CoRR, vol. abs/1806.08722}, 2018.

\bibitem{luo_et_al:OASIcs:2019:10188}
B.~Luo, J.~Shen, Y.~Wang, and M.~Pantic.
\newblock {The iBUG Eye Segmentation Dataset}.
\newblock In E.~Pirovano and E.~Graversen, editors, {\em 2018 Imperial College
  Computing Student Workshop (ICCSW 2018)}, volume~66 of {\em OpenAccess Series
  in Informatics (OASIcs)}, pages 7:1--7:9, Dagstuhl, Germany, 2019. Schloss
  Dagstuhl--Leibniz-Zentrum fuer Informatik.

\bibitem{mcmurrough2012eye}
C.~D. McMurrough, V.~Metsis, J.~Rich, and F.~Makedon.
\newblock An eye tracking dataset for point of gaze detection.
\newblock In {\em Proceedings of the Symposium on Eye Tracking Research and
  Applications}, pages 305--308. ACM, 2012.

\bibitem{paszke2017automatic}
A.~Paszke, S.~Gross, S.~Chintala, G.~Chanan, E.~Yang, Z.~DeVito, Z.~Lin,
  A.~Desmaison, L.~Antiga, and A.~Lerer.
\newblock Automatic differentiation in pytorch.
\newblock 2017.

\bibitem{Patney:2016:PFV:2929464.2929472}
A.~Patney, J.~Kim, M.~Salvi, A.~Kaplanyan, C.~Wyman, N.~Benty, A.~Lefohn, and
  D.~Luebke.
\newblock Perceptually-based foveated virtual reality.
\newblock In {\em ACM SIGGRAPH 2016 Emerging Technologies}, SIGGRAPH '16, pages
  17:1--17:2, New York, NY, USA, 2016. ACM.

\bibitem{peng2017large}
C.~Peng, X.~Zhang, G.~Yu, G.~Luo, and J.~Sun.
\newblock Large kernel matters--improve semantic segmentation by global
  convolutional network.
\newblock In {\em Proceedings of the IEEE conference on computer vision and
  pattern recognition}, pages 4353--4361, 2017.

\bibitem{proenca2010ubiris}
H.~Proenca, S.~Filipe, R.~Santos, J.~Oliveira, and L.~A. Alexandre.
\newblock The ubiris. v2: A database of visible wavelength iris images captured
  on-the-move and at-a-distance.
\newblock {\em IEEE Transactions on Pattern Analysis and Machine Intelligence},
  32(8):1529--1535, 2010.

\bibitem{ubiris2}
H.~{Proenca}, S.~{Filipe}, R.~{Santos}, J.~{Oliveira}, and L.~A. {Alexandre}.
\newblock The ubiris.v2: A database of visible wavelength iris images captured
  on-the-move and at-a-distance.
\newblock {\em IEEE Transactions on Pattern Analysis and Machine Intelligence},
  32(8):1529--1535, Aug 2010.

\bibitem{radu2015robust}
P.~Radu, J.~Ferryman, and P.~Wild.
\newblock A robust sclera segmentation algorithm.
\newblock In {\em 2015 IEEE 7th International Conference on Biometrics Theory,
  Applications and Systems (BTAS)}, pages 1--6. IEEE, 2015.

\bibitem{rot2018deep}
P.~Rot, {\v{Z}}.~Emer{\v{s}}i{\v{c}}, V.~Struc, and P.~Peer.
\newblock Deep multi-class eye segmentation for ocular biometrics.
\newblock In {\em 2018 IEEE International Work Conference on Bioinspired
  Intelligence (IWOBI)}, pages 1--8. IEEE, 2018.

\bibitem{sankowski2010reliable}
W.~Sankowski, K.~Grabowski, M.~Napieralska, M.~Zubert, and A.~Napieralski.
\newblock Reliable algorithm for iris segmentation in eye image.
\newblock {\em Image and vision computing}, 28(2):231--237, 2010.

\bibitem{DBLP:journals/corr/abs-1809-04729}
A.~Shafaei, M.~Schmidt, and J.~J. Little.
\newblock Does your model know the digit 6 is not a cat? {A} less biased
  evaluation of "outlier" detectors.
\newblock {\em CoRR}, abs/1809.04729, 2018.

\bibitem{shrivastava2017learning}
A.~Shrivastava, T.~Pfister, O.~Tuzel, J.~Susskind, W.~Wang, and R.~Webb.
\newblock Learning from simulated and unsupervised images through adversarial
  training.
\newblock In {\em Proceedings of the IEEE Conference on Computer Vision and
  Pattern Recognition}, pages 2107--2116, 2017.

\bibitem{Smith2013GazeLP}
B.~A. Smith, Q.~Yin, S.~K. Feiner, and S.~K. Nayar.
\newblock Gaze locking: passive eye contact detection for human-object
  interaction.
\newblock In {\em UIST}, 2013.

\bibitem{Tanriverdi2000}
V.~Tanriverdi and R.~J. Jacob.
\newblock Interacting with eye movements in virtual environments.
\newblock In {\em Proceedings of the SIGCHI conference on Human Factors in
  Computing Systems}, pages 265--272. ACM, 2000.

\bibitem{thoma2016survey}
M.~Thoma.
\newblock A survey of semantic segmentation.
\newblock {\em arXiv preprint arXiv:1602.06541}, 2016.

\bibitem{tonsen2016labelled}
M.~Tonsen, X.~Zhang, Y.~Sugano, and A.~Bulling.
\newblock Labelled pupils in the wild: a dataset for studying pupil detection
  in unconstrained environments.
\newblock In {\em Proceedings of the Ninth Biennial ACM Symposium on Eye
  Tracking Research \& Applications}, pages 139--142. ACM, 2016.

\bibitem{venkateswarlu2003eye}
R.~Venkateswarlu et~al.
\newblock Eye gaze estimation from a single image of one eye.
\newblock In {\em Proceedings Ninth IEEE International Conference on Computer
  Vision}, pages 136--143. IEEE, 2003.

\bibitem{wang2018hierarchical}
K.~Wang, R.~Zhao, and Q.~Ji.
\newblock A hierarchical generative model for eye image synthesis and eye gaze
  estimation.
\newblock In {\em Proceedings of the IEEE Conference on Computer Vision and
  Pattern Recognition}, pages 440--448, 2018.

\bibitem{wood2016learning}
E.~Wood, T.~Baltru{\v{s}}aitis, L.-P. Morency, P.~Robinson, and A.~Bulling.
\newblock Learning an appearance-based gaze estimator from one million
  synthesised images.
\newblock In {\em Proceedings of the Ninth Biennial ACM Symposium on Eye
  Tracking Research \& Applications}, pages 131--138. ACM, 2016.

\bibitem{wood2015rendering}
E.~Wood, T.~Baltrusaitis, X.~Zhang, Y.~Sugano, P.~Robinson, and A.~Bulling.
\newblock Rendering of eyes for eye-shape registration and gaze estimation.
\newblock In {\em Proceedings of the IEEE International Conference on Computer
  Vision}, pages 3756--3764, 2015.

\bibitem{Xiao:2018:DLI:3272127.3275032}
L.~Xiao, A.~Kaplanyan, A.~Fix, M.~Chapman, and D.~Lanman.
\newblock Deepfocus: Learned image synthesis for computational displays.
\newblock {\em ACM Trans. Graph.}, 37(6):200:1--200:13, Dec. 2018.

\bibitem{7299081}
X.~{Zhang}, Y.~{Sugano}, M.~{Fritz}, and A.~{Bulling}.
\newblock Appearance-based gaze estimation in the wild.
\newblock In {\em 2015 IEEE Conference on Computer Vision and Pattern
  Recognition (CVPR)}, pages 4511--4520, June 2015.

\bibitem{evalAI}
D.~{Deshraj}, R.~{Jain}, H.~{Agrawal}, P.~{Chattopadhyay}, T.~{Singh}, A.~{Jain}, B.~{Singh}, S.~{Lee} and D.~{Batra}.
\newblock EvalAI: Towards Better Evaluation Systems for AI Agents
\newblock In {\em arXiv},  1902.03570, 2019.
\end{thebibliography}
}

\end{document}